# Analysis on Riemann Hypothesis with Cross Entropy Optimization and Reasoning

Kevin Li, Fulu Li

Contact: fulu@alum.mit.edu

**Abstract**

In this paper, we present a novel framework for the analysis of Riemann Hypothesis [27], which is composed of three key components: a) probabilistic modeling with cross entropy optimization and reasoning; b) the application of the law of large numbers; c) the application of mathematical inductions. The analysis is mainly conducted by virtue of probabilistic modeling of cross entropy optimization and reasoning with rare event simulation techniques. The application of the law of large numbers [2, 3, 6] and the application of mathematical inductions make the analysis of Riemann Hypothesis self-contained and complete to make sure that the whole complex plane is covered as conjectured in Riemann Hypothesis. We discuss finite precision modeling in terms of the computation of the value of Riemann Zeta functions from a practical perspective with respect to the quantization modeling and the number of bits for the variables commonly used in today's computing facilities. We also discuss the method of enhanced top-*p* sampling with large language models (LLMs) for reasoning, where next token prediction is not just based on the estimated probabilities of each possible token in the current round but also based on accumulated path probabilities among multiple top-*k* chain of thoughts (CoTs) paths. Essentially, we propose the use of a combination of top-*p* sampling method and beam search with a beam width of *k* for chain of thoughts (CoTs) reasoning with large language models (LLMs). The probabilistic modeling of cross entropy optimization and reasoning may suit well with the analysis of Riemann Hypothesis as Riemann Zeta functions are inherently dealing with the sums of *infinite* components of a complex number series.

We hope that our analysis in this paper could shed some light on some of the insights of Riemann Hypothesis, in particular from the perspective of of probability convergence of random samples with cross entropy optimization and reasoning as well as the application of the law of large numbers and the application of mathematical inductions as well as the method of enhanced top-*p* sampling with large language models (LLMs) for reasoning. The framework and techniques presented in this paper, coupled with recent developments with chain of thought (CoT) or diagram of thought (DoT) reasoning in large language models (LLMs) with reinforcement learning (RL) [1, 7, 18, 21, 24, 34, 39-41], could pave the way for eventual proof of Riemann Hypothesis [27].

**Keywords:** Riemann Hypothesis, Randomness, Cross Entropy Optimization and Reasoning, Rare Event Simulation, Probabilistic Modeling, Law of Large Numbers, Mathematical Inductions, Complex Analysis, Chain of Thought (CoT), Reinforcement Learning (RL), Large Language Models (LLMs), Diagram of Thought (DoT)

## 1. Introduction

About 165 years ago, Riemann Hypothesis (RH) was proposed by Bernhard Riemann in 1859 regarding the input-value conditions, i.e., input variables, and the zero-function-value distributions of Riemann Zeta function, which could have great implications on the distribution of prime numbers [27] and other well-known mathematical conjectures such as Goldbach's conjecture for odd numbers [14]. The Riemann Zeta function or Euler-Riemann Zeta function was first introduced by Leonhard Euler in the early 1700 when Euler computed the *infinite* sum of the inverse of squares of integers. Euler studied the Zeta functions with the input variable of real numbers in the first half of the $18^{th}$ century. It was Bernhard Riemann who extended Euler's definition of the Zeta function to complex variables. Riemann conjectured in his landmark paper [27] that Riemann Zeta function has its complex zeros only at the negative even integers for the real part, where the imaginary part is zero, and complex numbers with real part of $\frac{1}{2}$, which is widely regarded as the critical line in this literature. For those zero function values of Riemann Zeta function of $\zeta(s)$ with $s$ at negative even integers such as -2, -4, -6, -8 …, they are called *trivial* zeros. Riemann Zeta function of $\zeta(s)$ is also zero for other values of $s$, other than those negative even integers, which are called *non-trivial* zeros. We need to emphasize that Riemann Hypothesis is only concerned with the locations of those *non-trivial* zeros. Riemann Hypothesis states that the real part of $s$, i.e., $\Re(s)$, for every *non-trivial* zero of Riemann Zeta function of $\zeta(s)$ is $\frac{1}{2}$, i.e., $\{s \in \mathbb{C} : \Re(s) = \frac{1}{2}\}$.

Riemann proposed his conjecture in his paper titled "On the Number of Primes Less Than a Given Magnitude" [27] and Riemann Hypothesis has deep connections with the distribution of prime numbers inherently at the very beginning. It is indeed a rare feat that Riemann hypothesis involves two distinct and remarkable areas in mathematics: the number theory, which is regarded as the Queen of Mathematics by Carl Friedrich Gauss, and the complex analysis, which has founding contributions from notable mathematicians such as Augustin-Louis Cauchy, Leonhard Euler, Carl Friedrich Gauss, Bernhard Riemann, and others.

Riemann Hypothesis is widely considered by many as one of the hardest unsolved problems in mathematics. Despite numerous attempts by some of the best minds across quite a few generations in history, Riemann Hypothesis remains unsolved till today.

Notably, large language models (LLMs) have made significant progress in the last few years with deep-learning neural networks, where the discovery has been made for the possibility to find out causal attention with Transformer models [38] with massive data sets and massive parallel computation with graphics processing units (GPUs) or tensor processing units (TPUs). With recent advancement of mathematical reasoning with large language models (LLMs) [1, 7, 18, 21, 24, 34, 39-41], in particular with chain of thought (CoT) reasoning in LLMs with reinforcement learning (RL) [18, 24, 39], great strides have been made possible for ultimate problem-solving of some seemingly-unsolvable mathematical problems such as the proof of Riemann Hypothesis.

In this paper, we present a novel framework for the analysis of Riemann Hypothesis, which is composed of three key components: a) probabilistic modeling with cross-entropy optimization and reasoning; b) the application of the law of large numbers; c) the application of mathematical inductions. The analysis is mainly conducted by virtue of probabilistic modeling of cross entropy optimization and reasoning with rare event simulation techniques. The application of the law of large numbers [2, 3, 6] and the application of mathematical inductions make the analysis of Riemann Hypothesis self-contained and complete. We discuss finite precision modeling in terms of the computation of the value of Riemann Zeta functions from a practical perspective with respect to the quantization modeling and the number of bits for the variables commonly used in the central processing units (CPUs) or graphics processing units (GPUs) or tensor processing units (TPUs) in today's computing facilities. We also discuss the method of enhanced top-$p$ sampling with large language models (LLMs) for reasoning, where next token prediction is not just based on the estimated probabilities of each possible token in the current round but also based on accumulated path probabilities among multiple top-$k$ paths. The probabilistic modeling of cross entropy optimization and reasoning may suit well with the analysis of Riemann Hypothesis as Riemann Zeta functions are inherently dealing with the sums of *infinite* components of a complex number series.

As discussed in [25] that the truth of Riemann Hypothesis might eventually depend on the techniques of a mixture of *analytic* and *arithmetic* ingredients, which is truly an inspiration to motivate the work in this paper, where probabilistic cross-entropy optimization and reasoning *analysis* of Riemann Hypothesis meets with random sampling of Riemann Zeta function with *arithmetic* computations.

As stated by the famous quote of Isaac Newton, if we have seen further, it is by standing on the shoulders of the giants. We sincerely solute each and every one of them who has made contributions in the pursuit of the proof of Riemann Hypothesis, in the fields of cross-entropy optimization and reasoning, mathematical reasoning with large language models (LLMs), the law of large numbers and mathematical inductions, and probabilistic modeling, etc.

The rest of the paper is organized as follows: We discuss the related work in Section 2. We present an in-depth analysis on Riemann Hypothesis with cross entropy optimization and reasoning in Section 3. We discuss the application of Bernoulli's law of large numbers in Section 4. We present the application of mathematical inductions in Section 5. We discuss the method of enhanced top-$p$ sampling method with LLMs for reasoning in Section 6. Summary and future directions are given in Section 7.

## 2. Related Work

**In the Pursuit of the Proof of Riemann Hypothesis**. In the early 1900, Hardy in [13] and Hardy, Fekete and Littlewood in [14] proved that there are *infinitely* many zeros for Riemann Zeta functions on the critical line, by considering moments of certain functions related to the Riemann Zeta function. Alan M.Turing, who is widely considered as the farther of Artificial Intelligence (AI), and Lehman presented methods for the computation of the number of zeros for Riemann Zeta functions in a given region [19, 37].

In general, the pursuit of the proof of Riemann Hypothesis can be classified into two categories: a) Transforming Riemann Hypothesis into another equivalent problem of $B$, by proving the correctness of problem $B$, essentially Riemann Hypothesis also gets proved; b) Computing and locating zeros of Riemann Zeta functions in the critical strip region of $\{s \in \mathbb{C} : \ 0 < \Re(s) < 1\}$ and verifying that if those non-trivial zeros are all on the critical line of $\{s \in \mathbb{C} : \ \Re(s) = \frac{1}{2}\}$.

For the first category, we can take a look at the following example: The de Bruijn–Newman constant, denoted by $\Lambda$, is defined as the unique real number such that the function $H(\lambda, z)$, defined with a real parameter $\lambda$, a complex variable $z$ and a super-exponentially decaying function of $\Phi(u)$, has only real zeros if and only if $\lambda \geq \Lambda$. Since Riemann Hypothesis is equivalent to the claim that all the zeroes of $H(0,z)$ are real, Brad Rodgers and Terence Tao discovered that Riemann Hypothesis is equivalent to $\Lambda = 0$ by proving zero to be the lower bound of the constant of $\Lambda$ [30]. Proving zero to be the upper bound of $\Lambda$ would also prove the correctness of Riemann hypothesis.

For the second category, we can take a look at the following endeavors: The basic idea is to verify Riemann Hypothesis up to a given imaginary part of $T$. For Riemann Zeta function of $\zeta(s)$ with a complex variable of $s$, we have $s = \sigma + it$, where $t$ is the imaginary part. This is done by computing the total number of zeros of Riemann Zeta function in a given region using Turing method [19,37] and checking that if it is the same as the number of zeros found on the critical line of $\{s \in \mathbb{C} : \ \Re(s) = \frac{1}{2}\}$, as conjectured in Riemann Hypothesis. As pointed out in [25] that extensive computations by O. Bohigas and M. Gianonni in [5], and by A. Odlyzko in [23] have confirmed the statistics with high accuracy for the verification of Riemann Hypothesis. In particular, Platt and Trudgian have reported that the computation of Riemann zeros with accuracy to various millions and imaginary parts to the height of $3 \times 10^{12}$ in the year of 2020 [26]. We refer interested readers on how numerical calculations of Riemann Zeta function is done with respect to the critical line to [5, 23, 26] for details.

In addition, there are quite a lot of efforts for better understanding of Riemann Hypothesis and Riemann Zeta functions during the last 165 years and we just list a few of them here due to limited space [4, 5, 8-17, 19, 22, 23, 25, 26, 30, 35-37], all of which constitute this seemingly never-ending tide of pursuit for the proof of Riemann Hypothesis, wherever it might lead to.

**Probabilistic Modeling with Cross Entropy Optimization and Reasoning**. The basic idea of cross entropy (CE) method [20, 31, 32] is to translate the deterministic optimization problem into a corresponding stochastic one and then use rare event simulation techniques to find the optimal solution. It works in an iterative fashion by generating a large number of random samples at each round based on some probability distribution over a given value set. At the end of each round, performance evaluation is conducted for each random sample with respect to the optimization objectives. The central step of cross entropy method is to adjust the probability distribution based on which the random samples are generated, where those samples with better performance in the last round in terms of the optimization objectives are given higher probabilities to generate samples in the current round.

In [20], Li et al presented a random tree optimization approach for the construction of the most parsimonious phylogenetic trees based on CE method, where an iterative optimization process is applied to reduce the parsimony score of a tree with the principle of cross entropy optimization. As discussed in [20, 31, 32], cross entropy (CE) method differs from other well-known random search algorithms for global optimization such as simulated annealing, tabu search and genetic algorithms, which are local search heuristics and employ the notion of local neighborhood structures. On the other hand, cross entropy (CE) method employs multi-extremal optimization process based on stochastic Kullback-Leibler cross-entropy, importance sampling, etc. Therefore, cross entropy (CE) method represents a *global* random search procedure rather than a *local* one. We refer interested readers to [31, 32] for the details of the proofs.

We will elaborate on the details of how we use cross entropy optimization and reasoning for the analysis of Riemann Hypothesis in Section 3.

**Mathematical Reasoning with Large Language Models (LLMs)**. In the year of 1953, Albert Einstein once commented on the development of Western Science and he concluded that "the development of Western Science in based on two great achievements: the invention of the formal logical system by Greek philosophers, and the discovery of the possibility to find out causal relationships by systematic experiments (Renaissance)". Large language models (LLMs) made great strides or breakthroughs with deep neural networks via unsupervised learning in recent years mainly due to the discovery of the possibility to find out causal attention by Transformer models [38] with massive data sets and massive parallel computations with graphics processing units (GPUs) or tensor processing units (TPUs).

With recent advancement of mathematical reasoning with large language models (LLMs) [1, 7, 18, 21, 24, 34, 39-41], in particular with chain of thought (CoT) reasoning in LLMs with reinforcement learning (RL) [18, 24, 39], the possibility is high for eventual tackling of some seemingly-unsolvable mathematical problems such as the proof of Riemann Hypothesis. In [21], Li et al gave an in-depth analysis on why the form of chain of thought (CoT) improves the reasoning capability of large language models (LLMs) on arithmetic and symbolic reasoning tasks for mathematical problem-solving. The authors in [1, 7, 18, 34, 39, 40] all emphasized the importance of scaling and optimizing compute time for inference and reasoning in large language models (LLMs) for mathematical reasoning tasks.

In [41], Zhang et al presented the framework of diagram of thought (DoT) for next-generation reasoning-specialized large language models (LLMs) with a directed acyclic graph (DAG) within a single model. The newly-released GPT-o1 by OpenAI in [24] demonstrated tremendous ways of learning to reason with large language models (LLMs) by means of chain of thought (CoT) and reinforcement learning (RL).

Our work differs from existing approaches in that we use probabilistic modeling with cross entropy optimization and reasoning in which a large number of random samples are generated based on some probability distributions for the computation of Riemann Zeta functions. Probability distributions over a given region for the computation of Riemann Zeta functions is updated after each round based on the characteristics of the values of Riemann Zeta function for those random samples, i.e., if the value of Riemann Zeta function of $\zeta(s)$ is zero and if the value of $s$ is on the critical line, i.e., $\Re(s) = \frac{1}{2}$. We will present the details on the analysis of Riemann Hypothesis with cross entropy optimization and reasoning in Section 3. We use the application of Bernoulli's law of large numbers and the application of mathematical inductions to make sure that the whole complex plane is covered for the computation of Riemann Zeta functions as conjectured in Riemann Hypothesis. We also discuss the method of enhanced top-$p$ sampling with large language models (LLMs) for reasoning, where next token prediction is not just based on the estimated probabilities of each possible token in the current round but also based on accumulated path probabilities among multiple top-$k$ paths.

## 3. Analysis of Riemann Hypothesis with Cross Entropy Optimization and Reasoning

**Riemann Zeta function revisited.** Riemann Zeta function, say $\zeta(s)$, is a function of a complex variable of $s$, where $s = \sigma + it$, with $\sigma$ and $t$ being real numbers. Riemann Zeta function can be written as a converging summation or as an integral when $\Re(s) = \sigma > 1$:

$$\zeta(s) = \sum_{n=1}^{\infty} \frac{1}{n^s} = \frac{1}{\Gamma(s)} \int_0^{\infty} \frac{x^{s-1}}{e^x - 1} dx, \qquad (1)$$

where

$$\Gamma(s) = \int_0^{\infty} x^{s-1} e^{-x} dx \qquad (2)$$

Riemann extended Riemann Zeta function beyond $\Re(s) = \sigma > 1$ to the whole complex plane by means of analytic continuation with Riemann's functional equation that he discovered. The basic idea is to first establish some functional equation on smaller domain, then using this functional equation to extend to the whole complex domain.

**Riemann's functional equation.** Riemann discovered that Riemann Zeta function satisfied the functional equation as follows:

$$\zeta(s) = 2^s \pi^{s-1} \sin(\frac{\pi s}{2}) \Gamma(1-s) \zeta(1-s), \qquad (3)$$

where $\Gamma(s)$ is the Gamma function indicated in Equation (2). Equation (3) is an equality of meromorphic functions that are valid on the whole complex plane. From Equation (3), we can observe that Riemann Zeta function $\zeta(s)$ has the value of zeros when $s = -2n$, where $s$ is even negative values due to the zero value of $\sin(\frac{\pi s}{2})$. However, when $s = 2n$, where $s$ is even positive values, the product of $\sin(\frac{\pi s}{2}) \Gamma(1-s)$ on the right side of the equation in Equation (3) is *not* zero. This is due to the fact that Gamma function of $\Gamma(1-s)$ has a simple pole, which cancels out the simple zero of the sine factor of $\sin(\frac{\pi s}{2})$.

Please note that for random sampling of Riemann Zeta function in our proposed cross-entropy optimization and reasoning in the critical strip of $\{s \in \mathbb{C} : 0 < \Re(s) < 1\}$, we use Riemann's functional equation, i.e., Equation (3), for arithmetic calculations of Riemann Zeta function value.

**Cross entropy optimization and reasoning.** The basic idea is that we examine the distribution of Riemann Zeta function zeros in the critical strip region of $\{s \in \mathbb{C} : 0 < \Re(s) < 1\}$ by random sampling of $s$ in a given sub-region of the critical strip region based on some probability distribution of $p(s)$ for the computation of Riemann Zeta function value of $\zeta(s)$ in an iterative process with cross entropy optimization and reasoning. The probability distribution of $p(s)$ is updated after each round based on some performance metric with respect to the distribution of Riemann Zeta function zeros in the critical strip region (see Equation (4) for details). If the number of random samples of $s$ for each round is large enough and the number of iterations in cross entropy optimization and reasoning is big enough, probability convergence can be observed with respect to the distribution of Riemann Zeta function zeros, which is the central point of Riemann Hypothesis. We use the application of Bernoulli's law of large numbers to guarantee that the observed distribution of Riemann Zeta function zeros is asymptomatically approaching its probability. The total number of Riemann Zeta function zeros can be computed for a given region according to Turing method [19, 37]. We use the application of mathematical inductions to make sure that the whole complex plane can be covered as conjectured in Riemann Hypothesis by organically expanding the examined sub-regions of the critical strip region to the whole critical strip region and beyond.

Following [21], we assume that the computation of Riemann Zeta function values are done with 32 or 64 bit floating point numbers, where the output of each arithmetic operation is rounded to the closest floating point number representable by a fixed number of digits following IEEE 754 standard, avoiding the unrealistic infinite precision assumptions during the calculation of the values of Riemann Zeta function of $\zeta(s)$ for $s \in \mathbb{C}$. We refer interested readers to [21] for details on the finite precision modeling for arithmetic calculations.

We have some basic notations for the analysis of Riemann Hypothesis with cross entropy optimization and reasoning as follows:. Let $f(\zeta(s))$ stand for the performance metric for a given Riemann Zeta function value with a given value of $s$ and we have the following definitions of $f(\zeta(s))$:

$$f(\zeta(s)) = \begin{cases} 1, & \text{if } \zeta(s) = 0 \text{ and } \Re(s) = \frac{1}{2} \\ -\infty, & \text{if } \zeta(s) = 0 \text{ and } \Re(s) \neq \frac{1}{2} \\ 0, & \text{otherwise} \end{cases} \qquad (4)$$

where $s$ is a complex number in the critical strip region, i.e., $\{s \in \mathbb{C} : 0 < \Re(s) < 1\}$.

Recall that Riemann Zeta function of $\zeta(s)$ is a function of a complex variable of $s$, where $s = \sigma + it$, with $\sigma$ and $t$ being real numbers. For a given sub-region of the critical strip region, where $\sigma \in (0, 1)$, $t \in [a, b]$ and $a, b$ are real numbers. Initially, to generate a random sample of $s$, we need to get a random number with a uniform distribution in a given range of $(0, 1)$ for real part number of $\sigma$ and a random number with a uniform distribution in a given range of $[a, b]$ for imaginary part number of $t$, respectively (see Figure 1).

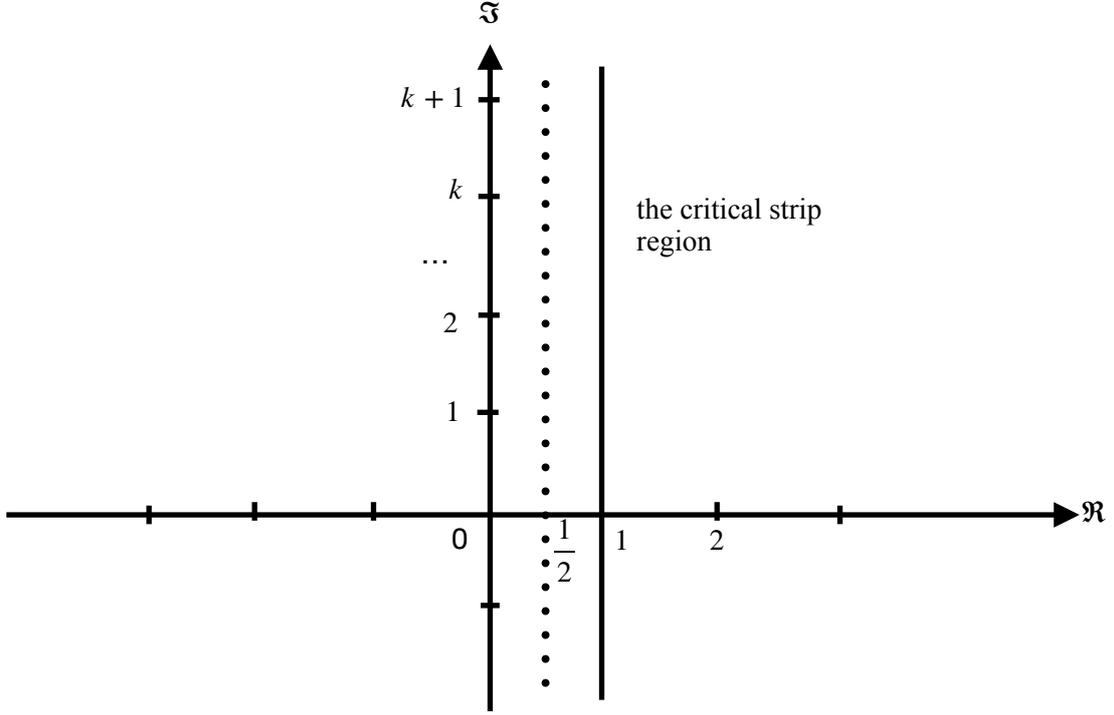

**Figure 1**: large number of random samples of $s$ for the calculation of Riemann Zeta function values of $\zeta(s)$ are drawn from the critical strip region of $\{s \in \mathbb{C} : 0 < \Re(s) < 1\}$.

Let $\gamma_i$ stand for the benchmark value of $f(\zeta(s))$ for the $i^{th}$ round. We have its definition as follows:
$$\gamma_i = \max\{f : \Pr_{S_{i-1}} (f(\zeta(s)) \geq f) \geq \rho\} \qquad (5)$$
where the purpose of the cross entropy optimization and reasoning for the analysis of Riemann Hypothesis is to maximize the sum of $f(\zeta(s))$ for a given number of random samples, i.e., to find out more chance where the value of Riemann Zeta function of $\zeta(s)$ is zero and $s$ is on the critical line of $\{s \in \mathbb{C} : \Re(s) = \frac{1}{2}\}$, $\rho$ normally takes a value of 0.01 so that the event of obtaining high performance is not too rare, the value of $s$ is randomly chosen based on the probability distribution of the $s$ values in a given region, $S_{i-1}$ represents the set of randomly generated values of $s$ in the $(i-1)^{th}$ round. Essentially, $\gamma_i$ is the top $\rho$-quantile of the performers of the randomly generated $s$ values for the calculation of Riemann Zeta function values in the $i^{th}$ round.

As discussed in [20, 31, 32], there are several ways to set the termination conditions. Normally, if for some $i \geq l$, say $l = 5$, and we have
$$\gamma_i = \gamma_{i-1} = \ldots = \gamma_{i-l} \qquad (6)$$
then we stop the cross entropy optimization and reasoning process.

Please also note that due to the definition of the performance metric function of $f(\zeta(s))$, if there is any case of $f(\zeta(s))$ being equal to minus infinity, where the value of Riemann Zeta function of $\zeta(s)$ is zero but $s$ is not on the critical line of $\{s \in \mathbb{C} : \Re(s) = \frac{1}{2}\}$, then the cross entropy optimization and reasoning process of Riemann Hypothesis terminates. More formally, we have
$$\sum_{k=1}^{M} f(\zeta(s_k)) < 0, \qquad (7)$$

where $M$ stand for the total number of random samples of $s$ for each round and $s_k$ is the $i^{th}$ randomly generated sample value of $s$.

Let $H_{\{\}}$ be an indicator function, $q_i$ is the probability that the given $i^{th}$ value of $s_i$ is being chosen and $q_i$ is initialized as zeros at the beginning. Recall that we have $s = \sigma + it$ and $\sigma \in (0, 1), t \in [a, b]$. We assume a uniform distribution of the values of $\sigma$ in a given range of $(0, 1)$. We also assume a uniform distribution of the values of $t$ in a given range of $[a, b]$, where $b > a$ and $a, b \in \Re$. Randomly choosing a value of $\sigma$ in a given range of $(0, 1)$ with a uniform distribution is like generating a random number in the range of $(0, 1)$ with a uniform distribution. Randomly choosing a value of $t$ in a given range of $[a, b]$ with a uniform distribution is also like generating a random number in the range of $[a, b]$ with a uniform distribution. The updated value of $q_i$ can be estimated as:

$$q_i^e = \frac{H_{\{f(\zeta(s_i)) \geq \gamma \text{ and } f(\zeta(s_i)) \text{ in top } \rho-\text{quantile performers}\}}}{\sum_{k=1}^{M} H_{\{f(\zeta(s_k)) \geq \gamma \text{ and } f(\zeta(s_k)) \text{ in top } \rho-\text{quantile performers}\}}} \quad (8)$$

While there are solid theoretical justifications for Equation (8), we refer interested readers to [31, 32] for the details. Please note that we added the condition of "$f(\zeta(s_i))$ in top $\rho$-quantile performers" of the randomly generated values of $s$ for the calculation of Riemann Zeta function values in the given round due to the fact that the threshold of $\gamma$ may not be accurate enough due to the nature of the discrete function value definition of $f(\zeta(s))$ in Equation (4) in particular at the early stages of the cross entropy optimization and reasoning process for the analysis of Riemann Hypothesis.

In order to have a smoothed update procedure, normally we have

$$q_i^m = c \times q_i^e + (1 - c) \times q_i^{m-1} \quad (9)$$

where empirically a value of $c$ between 0.4 and 0.9, i.e., $0.4 \leq c \leq 0.9$, gives the best results [31, 32], $q_i^{m-1}$ is the value of $q_i$ in the previous round, $q_i^e$ is the estimated value of $q_i$ based on the performance in the previous round according to Equation (8), $q_i^m$ stands for the value of $q_i$ in the current round.

Let $B$ be the set of values of $s$ whose performance are in the top $\rho$-quantile of the performers of the randomly generated values of $s$ for the calculation of Riemann Zeta function values based on the performance metrics defined in Equation (4) in a given round and we have

$$B = \{s_i | H_{\{f(\zeta(s_i)) \geq \gamma \text{ and } f(\zeta(s_i)) \text{ in top } \rho-\text{quantile performers}\}}\} \quad (10)$$

Let $q_i^n$ indicate the normalized $q_i^m$ among the top $\rho$-quantile of the performers and we have:

$$q_i^n = \frac{q_i^m}{\sum_{s_j \in B} q_j^m} \quad (11)$$

Let $v$ be the factor of favorability, i.e., $v = 10$, towards those top $\rho$-quantile of the performers in a given round and let $N_v$ be the total number of samples chosen from those top $\rho$-quantile of the performers for the next round and we have

$$N_v = v \times M \times \rho \quad (12)$$

For the next round, we have $N_v$ samples that are randomly chosen from those $|B|$ number of values of $s$ based on their normalized probability of $q_i^n$ (see Equation (11)), where $|B|$ indicates the number of elements in the value set of $B$. The rest of the samples, i.e., $(M - N_v)$ samples for $s = \sigma + it$, are randomly chosen from the ranges of $[a, b]$ for $t$ and $(0, 1)$ for $\sigma$ with uniform distributions, which can be accomplished by generating $(M - N_s)$ random numbers in the range of $[a, b]$ for $t$ and in the range of $(0, 1)$ for $\sigma$ with uniform distributions.

By choosing proper values of the hyper-parameters of $M$, $v$, $\rho$, $a$, $b$, $c$, $l$ and others, as the iteration process of cross entropy optimization and reasoning for the analysis of Riemann Hypothesis goes on, the cross entropy optimization and reasoning proceeds as follows: if there is any case of $\zeta(s) = 0$ and $\{s \in \mathbb{C} : \Re(s) \neq \frac{1}{2}\}$, i.e., Riemann Zeta function zero is not located on the critical line for the value of $s = \sigma + it$, then Riemann Hypothesis does not hold. According to the performance metrics that we defined in Equation (4), if $\sum_{k=1}^{M} f(\zeta(s_k)) < 0$, then Riemann Hypothesis does not hold. Otherwise, more and more cases of $\zeta(s) = 0$ and $\{s \in \mathbb{C}: \Re(s) = \frac{1}{2}\}$, i.e., Riemann Zeta function zeros being located on the critical line as conjectured in Riemann Hypothesis appear in the top $\rho$-quantile of the performers among the very large number of random samples of $s$ in the calculation of the values of Riemann Zeta function. After a large number of of iterations, probability convergence for those top $\rho$-quantile of the performers in terms of the values of $s$ will be achieved, where almost all of the top $\rho$-quantile of the performers among the very large number of random samples of $s$ will be the case of $\zeta(s) = 0$ and $\{s \in \mathbb{C}: \Re(s) = \frac{1}{2}\}$, i.e., Riemann Zeta function zeros being located on the critical line. The interesting reasoning dynamics is that more and more cases of $\zeta(s) = 0$ and $\{s \in \mathbb{C}: \Re(s) = \frac{1}{2}\}$ will appear in the top $\rho$-quantile of the performers while there are large enough room of random samples for the case of $\zeta(s) = 0$ and $\{s \in \mathbb{C}: \Re(s) \neq \frac{1}{2}\}$ to appear if there is any at each round.

In summary, for a given sub-region in the critical strip region of $\{s \in \mathbb{C}: 0 < \Re(s) < 1\}$, we have the algorithm of cross entropy optimization and reasoning (CEOR) for the analysis of Riemann Hypothesis as follows:

1. Set the round number variable $r = 1$ and initialize $q_i$ as zeros.
2. Generating $M$ random numbers in the range of $[a, b]$ for $t$ and in the range of $(0, 1)$ for $\sigma$ with uniform distributions for the values of $s = \sigma + it$ if it is the first time.
3. Generating $(M - N_v)$ random numbers in the range of $[a, b]$ for $t$ and in the range of $(0, 1)$ for $\sigma$ with uniform distribution for the values of $s = \sigma + it$ if it is not the first time.
4. $N_v$ is determined by Equation (12).
5. $N_v$ random samples of $s$ are drawn from those top $\rho$-quantile performers of $s$ in the last round based on probability distribution of $q_i^n$s
6. Calculate $\gamma_r$ according to Equation (5).
7. Update $q_i$ according to Equation (8) and Equation (9).
8. Update normalized $q_i$ according to Equation (11).
9. If $\sum_{k=1}^{M} f(\zeta(s_k)) < 0$, then stop.
10. If for some $r \geq l$, say $l = 5$, such that $\gamma_r = \gamma_{r-1} = \ldots = \gamma_{r-l}$, then stop, otherwise, reiterate from Step 3.

**Figure 2**: the CEOR algorithm for a given sub-region in the critical strip region of $\{s \in \mathbb{C}: 0 < \Re(s) < 1\}$ for the analysis of Riemann Hypothesis.

## 4. Application of Bernoulli's Law of Large Numbers

We use the application of Bernoulli's law of large numbers [2, 3, 6] to guarantee that when the number of random samples used in the cross entropy optimization and reasoning for the analysis of Riemann Hypothesis is large enough, the frequency of the event that the value of Riemann Zeta function of $\zeta(s)$ is zero and the corresponding $s$ is *not* located on the critical line of $\{s \in \mathbb{C}: \Re(s) = \frac{1}{2}\}$ is essentially asymptotically approaching to the probability of the event, where the total number of Riemann Zeta function zeros can be computed and the total number of zeros of Riemann Zeta functions for a given region is fixed [19, 37].

Let $\mu$ be the number of times that event $A$ occurs during $n$ independent trials and let $p$ denote the probability that event $A$ occurs for every trial. According to Bernoulli's law of large numbers [2, 3, 6], for any positive real number of $\epsilon$, we have

$$\lim_{n \to \infty} \Pr\{|\frac{\mu}{n} - p| < \epsilon\} = 1 \qquad (13)$$

In the case of the analysis of Riemann Hypothesis with cross entropy optimization and reasoning, event *A* is the event that the value of Riemann Zeta function of $\zeta(s)$ is zero and the corresponding *s* is *not* located on the critical line of $\{s \in \mathbb{C} : \Re(s) = \frac{1}{2}\}$. The random sampling for the calculation of Riemann Zeta function in a given region can be described by the binomial distribution: either the value of Riemann Zeta function of $\zeta(s)$ is zero and the corresponding *s* is *not* located on the critical line of $\{s \in \mathbb{C} : \Re(s) = \frac{1}{2}\}$ or not. When a given region is fixed, the total number of Riemann Zeta function zeros can be computed and the total number of zeros of Riemann Zeta functions is fixed according to Turing method [19, 37]. So, for a given region, the probability of *p* for the event that the value of Riemann Zeta function of $\zeta(s)$ is zero and the corresponding *s* is *not* located on the critical line of $\{s \in \mathbb{C} : \Re(s) = \frac{1}{2}\}$ exists and it is fixed. Therefore, Bernoulli's law of large numbers can be applied in this case.

## 5. Application of Mathematical Inductions

We use the application of mathematical indications to make sure that the whole complex plane is covered as conjectured in Riemann Hypothesis. For the sake of simplicity, let us first start from the critical strip region of $\{s \in \mathbb{C} : 0 < \Re(s) < 1\}$. The basic idea is to split the critical strip region into a series of contiguous smaller unit square regions, then we apply the techniques of mathematical inductions to prove that the whole complex plane can be covered for the computation of Riemann Zeta functions and for the analysis of Riemann Hypothesis. The application of mathematical inductions is needed in our analysis on Riemann Hypothesis with cross entropy optimization and reasoning as random samples of *s* for the computation of Riemann Zeta functions of $\zeta(s)$ with cross entropy optimization and reasoning are drawn from a fixed region and the whole complex plane has to be covered as conjectured in Riemann Hypothesis.

According to Turing method in [19, 37], the total number of zeros, i.e., the total number of cases where the value of Riemann Zeta function of $\zeta(s)$ is zero, can be computed for Riemann Zeta function of $\zeta(s)$ for a given region. Using the critical strip region of $\{s \in \mathbb{C} : 0 < \Re(s) < 1\}$ as an example, with the application of mathematical inductions, we show that the total number of zeros of Riemann Zeta functions can be computed for any given region across the whole complex plane and the total number of zeros of Riemann Zeta functions for any given region across the whole complex plane is fixed.

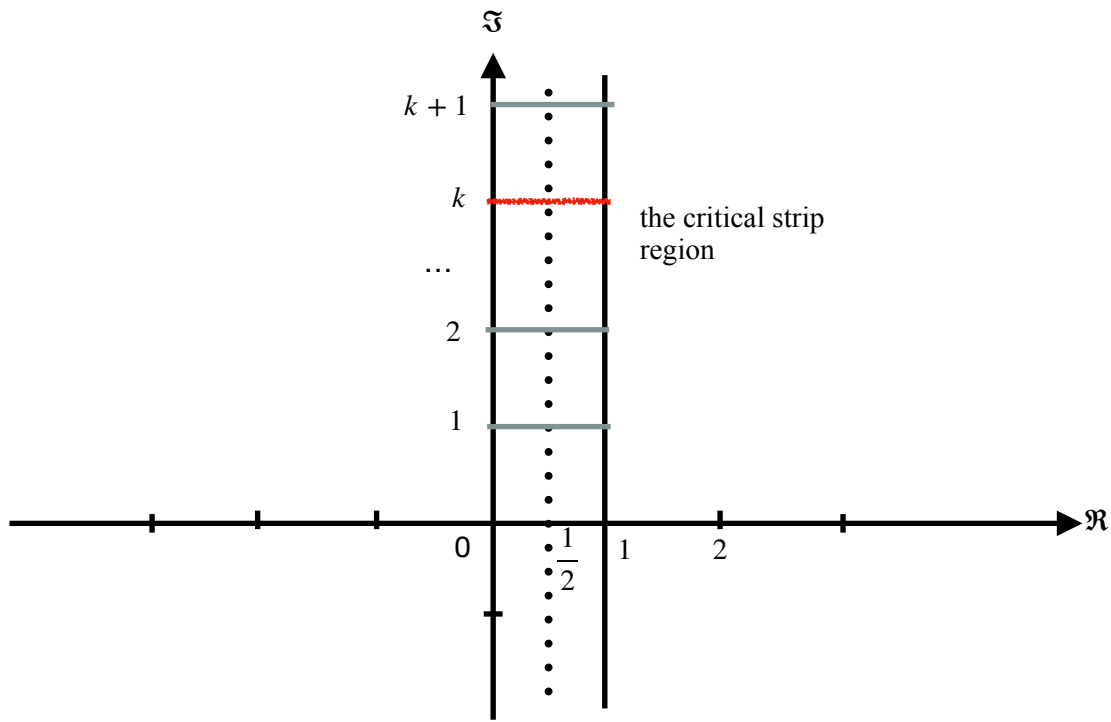

**Figure 3**: the application of mathematical inductions with continuous unit square regions in the critical strip region of $\{s \in \mathbb{C} : 0 < \Re(s) < 1\}$ for the calculation of Riemann Zeta function values of $\zeta(s)$.

Following the logic of mathematical inductions, let $n$ denote the number of continuous unit square regions, when $n$ is equal to 1, i.e., $n = 1$, for any unit square region in the critical strip region, i.e., $\{s \in \mathbb{C} : 0 < \Re(s) < 1\}$, $\{s \in \mathbb{C} : a < \Im(s) < (a+1)\}$ and $a$ is a real number, the conditions hold due to Turing method [19, 37], with which we can calculate the total number of Riemann Zeta function zeros for a given region and the total number of zeros of Riemann Zeta function is fixed for the given region.

As depicted in Figure 3, assuming that when $n$ is equal to $k$, for any $k$ continuous unit square regions, i.e., $\{s \in \mathbb{C} : 0 < \Re(s) < 1\}$, $\{s \in \mathbb{C} : a < \Im(s) < (a+k)\}$ and $a$ is a real number, the conditions hold where the total number of Riemann Zeta function zeros can be computed and the total number of zeros of Riemann Zeta functions is fixed. When $n$ is equal to $k + 1$, let us assume that for one specific $k + 1$ continuous unit square region, say, $\{s \in \mathbb{C} : 0 < \Re(s) < 1\}$, $\{s \in \mathbb{C} : b < \Im(s) < (b+k+1)\}$ and $b$ is a real number, the condition does not hold. Then, we can pick one region with $k$ continuous unit square regions, say, $\{s \in \mathbb{C} : 0 < \Re(s) < 1\}$, $\{s \in \mathbb{C} : b < \Im(s) < (b+k)\}$ and $b$ is the same real number as in that specific $k + 1$ continuous unit square region of $\{s \in \mathbb{C} : b < \Im(s) < (b+k+1)\}$, which contradicts with the assumption that for any $k$ continuous unit square regions, i.e., $\{s \in \mathbb{C} : 0 < \Re(s) < 1\}$, $\{s \in \mathbb{C} : a < \Im(s) < (a+k)\}$ and $a$ is a real number, the conditions hold where the total number of Riemann Zeta functions zeros can be computed and the total number of zeros of Riemann Zeta functions is fixed. Therefore, When $n$ is equal to $k + 1$, the conditions also hold where the total number of Riemann Zeta function zeros can be computed and the total number of zeros of Riemann Zeta functions is fixed. This concludes the proof.

By the application of mathematical inductions method, the *infinite* space over the whole complex plane for Riemann Hypothesis can be covered by continuous expansions of adding one or more continuous unit square regions for every expansion until the total coverage of the whole complex plane is achieved. In other words, the analysis of Riemann Hypothesis with cross entropy optimization and reasoning can be applied to the whole complex plane.

## 6. Enhanced Top-$p$ Sampling with LLMs for Reasoning

Mathematical reasoning capabilities of large language models (LLMs) can be improved with chain of thought (CoT) annotations and reinforcement learning [1, 7, 18, 21, 24, 34, 39-41]. In general, there are mainly two stages in which we can make further enhancement by adding diversity, i.e., more degree of freedom, in terms of either training paths or inference paths for large language models (LLMs). In the training stage, multiple reasoning paths with multiple valid chain of thought (CoT) annotations for the same question can be explored. On the other hand, for the inference stage, the chain of thought (CoT) generation process can be decomposed into a sequence of next token prediction actions and multiple chain of thoughts (CoTs) can be generated for the same question during the *intermediate* reasoning process to further enhance the reasoning capabilities of large language models (LLMs). In this paper, we only focus on the inference stage. Let $e$ stand for a chain of thought (CoT) process with a sequence of next token prediction actions of $a_1, a_2, \ldots, a_r$, the accumulated path probabilities of $p_e$ can be expressed as:

$$p_e = p_{a_1} \times p_{a_2} \times \ldots \times p_{a_r}, \tag{14}$$

where $p_{a_i}$ is the next token probability at the $i^{th}$ step of next token prediction action.

Notably, top-$p$ sampling method is widely used for next token prediction in Transformer-based [38] large language models due to its superior performance over greedy search, beam search, top-$k$ sampling (choosing $k$ most likely tokens) and others. Here is how it works: in top-$p$ sampling method, it chooses from the smallest possible set of tokens whose cumulative probability exceeds the probability $p$. The probability mass is then redistributed among this set of tokens by means of normalization. This way, the number of chosen tokens can dynamically change based on the next token's probability distribution.

In the following, we present the method of enhanced top-$p$ sampling with large language models (LLMs) for reasoning, in particular for the generation of the action sequence in multiple chain of thoughts (CoTs), where next token prediction is *not just* based on the estimated probabilities of each possible token in the current round *but also* based on accumulated path probabilities, i.e., chain of thought probabilities, among multiple top-$k$ paths as only $k$ chain of thoughts (CoTs) paths are kept during the intermediate reasoning process. Recall that for a decoder-only Transformer-based [38] large language model (LLM) system, a language model trained for causal language modeling takes into a sequence of text tokens as input and outputs the probability distribution for the next token, which is often expressed as logit in machine learning systems. The logit is also called the log-odds as it is equal to the *logarithm* of the odds $\frac{p}{1-p}$, where $p$ is a probability. Thus, the logit is a type of function that maps probability values from $(0,1)$ to real numbers in $(-\infty, +\infty)$. For the sake of simplicity, we use the notion of probability instead of logit in the following discussions.

As we discussed before, multiple chain of thoughts (CoT) action sequences can be generated during *intermediate* reasoning process. Next token prediction is *not just* based on the estimated probabilities of each possible token in the current round with top-*p* sampling method *but also* based on accumulated path probabilities, i.e., chain of thought probabilities, among multiple top-*k* paths as only *k* chain of thoughts (CoTs) paths are kept during the intermediate reasoning process. Essentially, we propose the use of a combination of top-*p* sampling method and beam search with a beam width of *k* for chain of thoughts (CoTs) reasoning with large language models (LLMs). By looking back at multiple chain of thoughts (CoTs) paths, back-tracking for *self-correction* is possible to avoid "making up" things for "*hallucinations*", which is one of the main challenges for current large language model (LLM) systems.

We also need to emphasize that it is possible to use *multiple* internal models for chain of thoughts (CoTs) reasoning, where top-*p* sampling method and accumulated path probabilities are used for generating chain of thoughts (CoTs) next token prediction actions and only *k* chain of thoughts (CoTs) paths are kept during the intermediate reasoning process. In the case of multiple internal models for chain of thoughts (CoTs) reasoning, at least one chain of thought (CoT) path has to be given for each internal model in order to maximize the benefits of reasoning with multiple internal models.

## 7. Summary and Future Directions

It has been 165 years since Bernhard Riemann presented his famed Riemann Hypothesis in his landmark paper [27]. Despite numerous notable attempts by some of the best minds in recent history, Riemann Hypothesis remains unsolved till today. As discussed in [25] that the truth of Riemann Hypothesis might eventually depend on the techniques of a mixture of *analytic* and *arithmetic* ingredients. In this paper, we present a novel framework for the analysis of Riemann Hypothesis, which is composed of three key components: a) probabilistic modeling with cross entropy optimization and reasoning; b) the application of the law of large numbers; c) the application of mathematical inductions. The analysis is mainly conducted by virtue of probabilistic modeling of cross entropy optimization and reasoning with rare event simulation techniques. The application of the law of large numbers [2, 3, 6] and the application of mathematical inductions make the analysis of Riemann Hypothesis self-contained and complete to make sure that the whole complex plane is covered as conjectured in Riemann Hypothesis. We discuss finite precision modeling in terms of the computation of the value of Riemann Zeta functions from a practical perspective with respect to the quantization modeling and the number of bits for the variables commonly used in today's computing facilities. We also discuss the method of enhanced top-*p* sampling with large language models (LLMs) for reasoning, where next token prediction is not just based on the estimated probabilities of each possible token in the current round but also based on accumulated path probabilities among multiple top-*k* chain of thoughts (CoTs) paths. Essentially, we propose the use of a combination of top-*p* sampling method and beam search with a beam width of *k* for chain of thoughts (CoTs) reasoning with large language models (LLMs). The probabilistic modeling of cross entropy optimization and reasoning may suit well with the analysis of Riemann Hypothesis as Riemann Zeta functions are inherently dealing with the sums of *infinite* components of a complex number series.

We will conduct extensive simulations for the analysis of Riemann Hypothesis with cross entropy optimization and reasoning for the validation of the distribution of Riemann Zeta function zeros as our future directions. We will also conduct extensive experiments to explore multiple chain of thoughts (CoTs) paths for the use of a combination of top-*p* sampling method in next token prediction actions and beam search with a beam width of *k* for chain of thoughts (CoTs) reasoning with large language models (LLMs) during intermediate reasoning process as one of our future endeavors.

The framework of cross entropy optimization and reasoning for the analysis of Riemann Hypothesis presented in this paper, coupled with recent developments with chain of thought (CoT) or diagram of thought (DoT) reasoning in large language models (LLMs) with reinforcement learning (RL) [1, 7, 18, 21, 24, 34, 39-41], could pave the way for eventual proof of Riemann Hypothesis [27].

# References

1. H. Bansal, A. Hosseini, R. Agarwal, V. Q. Tran, M. Kazemi, "Smaller, Weaker, Yet Better: Training LLM Reasoners via Compute-Optimal Sampling", https://arxiv.org/pdf/2408.16737, 2024.
2. J. Bernoulli, "Wahrscheinlichkeitsrechnung (Ars conjectandi, 1713)", Ostwalds Klassiker der exakten Wissenschaften, W. Engelmann, Leipzig, 1899.
3. N. Bernoulli, "Usu Artis conjectandi in Jure", Doctoral Thesis, Basel 1709. In: Die Werke von Jakob Bernoulli, Vol. 3, Birkhäuser Basel 1975.
4. M. Bischoff, "The Biggest Problem in Mathematics Is Finally a Step Closer to Being Solved", https://www.scientificamerican.com/article/the-riemann-hypothesis-the-biggest-problem-in-mathematics-is-a-step-closer/, July 2024.
5. O. Bohigas, M. Gianonni, "Chaotic motion and random matrix theories", Lecture Notes in Physics, 209, 1984.
6. E. Bolthausen, M. V. Wuthrich, "Bernoulli's Law of Large Numbers", https://people.math.ethz.ch/~wueth/Positions/2013_Bernoulli.pdf, 2013.
7. B. Brown, J. Juravsky, R. Ehrlich, R. Clark, Q. V. Le, C. Re, A. Mirhoseini, "Large Language Monkeys: Scaling Inference Compute with Repeated Sampling", https://arxiv.org/pdf/2407.21787, 2024.
8. D. M. Burton, "Elementary Number Theory", Tata McGraw-Hill Publishing Company Limited, ISBN 978-0-07-061607-3, 2006.
9. J. Cepelewicz, "'Sensational' Proof Delivers New Insights Into Prime Numbers", https://www.quantamagazine.org/sensational-proof-delivers-new-insights-into-prime-numbers-20240715/, July 2024.
10. J. B. Conrey, A. Ghosh and S. M. Gonek, "Simple Zeros of the Riemann Zeta Function", https://people.math.rochester.edu/faculty/gonek/papers/SimpleZeros.pdf, 1997.
11. H. M. Edwards, Riemann's zeta function. Pure and Applied Mathematics. Vol. 58. New York-London: Academic Press. ISBN 0-12-232750-0. Zbl 0315.10035, 1974.
12. L. Guth, J. Maynard, "New Large Value Estimates for Dirichlet Polynomials", https://arxiv.org/abs/2405.20552, May 2024.
13. G. H. Hardy, "Sur les zeros de la fonction ζ(s)". Comptes rendus de l'Académie des Sciences. 158. French Academy of Sciences: 1012–1014, 1914.
14. G. H. Hardy, M. Fekete, J.E. Littlewood, "The Zeros of Riemann's Zeta-Function on the Critical Line". Journal of the London Mathematical Society. s1-1: 15–19. doi:10.1112/jlms/s1-1.1.15, 1921.
15. W. L. Hosch, "Riemann hypothesis", https://www.britannica.com/science/Riemann-hypothesis.
16. A. Ivić, "The theory of Hardy's Z-function", Cambridge Tracts in Mathematics. Vol. 196. Cambridge: Cambridge University Press. ISBN 978-1-107-02883-8. Zbl 1269.11075, 2013.
17. V. Kargin, "Statistical properties of zeta functions' zeros", https://arxiv.org/pdf/1302.1452, Probability Surveys, 2014.
18. A. Kumar, V. Zhuang, R. Agarwal, Y. Su, et al, "Training Language Models to Self-Correct via Reinforcement Learning", https://arxiv.org/pdf/2409.12917, 2024.
19. R.S. Lehman, "On the Distribution of Zeros of the Riemann Zeta-Function". Proceedings of the London Mathematical Society. s3-20 (2): 303–320. doi:10.1112/plms/s3-20.2.303, 1970.
20. F. Li, and A. Lippman, "Random tree optimization for the construction of the most parsimonious phylogenetic trees", 43rd Annual Conference on Information Sciences and Systems, 2009.
21. Z. Li, H. Liu, D. Zhou, T. Ma, "Chain of Thought Empowers Transformers to Solve Inherently Serial Problems", https://arxiv.org/pdf/2402.12875, 2024.
22. Y. Lin, "Zeros of Riemann zeta function", https://math.uchicago.edu/~may/REU2019/REUPapers/Lin,Yuxin.pdf, 2019.
23. A. Odlyzko, On the distribution of spacings between zeros of the zeta functions, Math. Comp., 48, 1987.
24. OpenAI, "Learning to reason with LLMs", https://openai.com/index/learning-to-reason-with-llms/, 2024.
25. R. Perez-Marco, "Notes on Riemann hypothesis", https://arxiv.org/pdf/1707.01770, February 2018.
26. D. Platt, T. Trudgian, "The Riemann hypothesis is true up to $3 \times 10^{12}$", Bulletin of the London Mathematical Society, 53 (3), Wiley: 792–797, arXiv:2004.09765, doi:10.1112/blms.12460, S2CID 234355998, 2021.
27. B. Riemann, "On the Number of Prime Numbers Less Than a Given Magnitude", Monthly Reports of the Berlin Academy, 1859.
28. Riemann Hypothesis, https://www.claymath.org/library/annual_report/ar2004/04report_riemann.pdf, 2004.

29. https://mathworld.wolfram.com/RiemannHypothesis.html.
30. B. Rodgers, T. Tao, "The de Bruijn–Newman constant is non-negative", Forum of Mathematics, 8: e6, 62, arXiv:1801.05914, 2020.
31. R. Rubinstein, "The Cross-Entropy Method for Combinatorial and Continuous Optimization", Methodology And Computing in Applied Probability, 1999.
32. R. Rubinstein, D. Kroese, "The Cross-Entropy Method: A Unified Approach to Combinatorial Optimization, Monte-Carlo Simulation and Machine Learning", Springer, 2004.
33. P. Sarnak, "Problems of the Millennium: The Riemann Hypothesis", http://www.claymath.org/library/annual_report/ar2004/04report_sarnak.pdf, 2004.
34. C. Snell, J. Lee, K. Xu, A. Kumar, "Scaling LLM Test-Time Compute Optimally can be More Effective than Scaling Model Parameters", https://arxiv.org/pdf/2408.03314, 2024.
35. A. Sutherland, "Riemann's zeta function and the prime number theorem", https://math.mit.edu/classes/18.785/2021fa/LectureNotes16.pdf, 2021.
36. T. S. Trudgian, "An improved upper bound for the argument of the Riemann zeta function on the critical line II". J. Number Theory. 134: 280–292. arXiv:1208.5846. doi:10.1016/j.jnt.2013.07.017, 2014.
37. A. M. Turing, "Some Calculations of the Riemann Zeta-Function". Proceedings of the London Mathematical Society. s3-3 (1): 99–117. doi:10.1112/plms/s3-3.1.99, 1953.
38. A. Vaswani, N. Shazeer, N. Parmar, J. Uszkoreit, L. Jones, A. Gomez, L. Kaiser, I. Polosukhin, "Attention Is All You Need", https://arxiv.org/pdf/1706.03762.pdf, 2017.
39. J. Wei, X. Wang, D. Schuurmans, M. Bosma, B. Echter, F. Xia, E. Chi, Q. Le, D. Zhou, "Chain-of-Thought Prompting Elicits Reasoning in Large Language Models", https://arxiv.org/pdf/2201.11903, 2024.
40. Y. Wu, Z. Sun, S. Li, S. Welleck, Y. Yang, "An Empirical Analysis of Compute-Optimal Inference for Problem-Solving with Language Models", https://arxiv.org/pdf/2408.00724, 2024.
41. Y. Zhang, Y. Yuan, A. Yao, "On the Diagram of Thought", https://arxiv.org/pdf/2409.10038v1, 2024.